\pdfoutput=1

\documentclass{article}

\usepackage{arxiv}
\usepackage[utf8]{inputenc}
\usepackage[T1]{fontenc}
\usepackage{hyperref}
\hypersetup{
    colorlinks=true,
    linkcolor=blue,
    filecolor=magenta,      
    urlcolor=cyan,
}
\usepackage{url}
\usepackage{booktabs}
\usepackage{amsfonts}
\usepackage{nicefrac}
\usepackage{microtype}
\usepackage{lipsum}
\usepackage{graphicx}
\usepackage{subcaption}

\title{GACNN: Training Deep Convolutional Neural Networks with Genetic Algorithm}

\author{
  Parsa~Esfahanian\\
  Department of Computer Science\\
  Institute for Research in Fundamental Sciences\\
  \texttt{parsa.esfahanian@ipm.ir} \\
  \And
 Mohammad~Akhavan \\
  Department of Computer Science\\
 Institute for Research in Fundamental Sciences\\
  \texttt{mohammad.akhavan@ipm.ir} \\
}

\begin{document}
\maketitle

\begin{abstract}
Convolutional Neural Networks (CNNs) have gained a significant attraction in the recent years due to their increasing real-world applications. Their performance is highly dependent to the network structure and the selected optimization method for tuning the network parameters. 
In this paper, we propose novel yet efficient methods for training convolutional neural networks. The most of current state of the art learning method for CNNs are based on Gradient decent. In contrary to the traditional CNN training methods, we propose to optimize the CNNs using methods based on Genetic Algorithms (GAs). These methods are carried out using three individual GA schemes, Steady-State, Generational, and Elitism. We present new genetic operators for crossover, mutation and also an innovative encoding paradigm of CNNs to chromosomes aiming to  reduce the resulting chromosome's size by a large factor. We  compare the effectiveness and scalability of our encoding with the traditional encoding. Furthermore, the performance of individual GA schemes used for training the networks were compared with each other in means of convergence rate and overall accuracy. Finally, our new encoding alongside the superior GA-based training scheme is compared to Backpropagation training with Adam optimization.
\end{abstract}


\section{Introduction}
Recent years has seen the rapid growth of machine learning applications to the real world problems, specially deep learning. \cite{schmidhuber2015deep} Convolutional Neural Networks (CNNs), being one of the many classes of deep learning algorithms, are proving to be one of the most effective and popular tools used in fields such as computer vision and speech recognition.\\
Convolutional neural networks, with their exceptional generality in finding good solutions and a property to tolerate noisy and uncertain data, are becoming the go-to candidate for solving many problems. In addition, implementation, modification, testing, and application of CNNs for working on large sized datasets are becoming easier and more convenient, thanks to higher-end programming languages, like Python, and helpful libraries, such as Keras. \cite{chollet2015keras}\\
CNNs themselves are a special class of feedforward Artificial Neural Networks (ANNs), which are biologically inspired computation systems composed of simple processing elements, called neurons (or nodes), that are positioned in a layered fashion and interact with each other using weighted connections. A CNN's architecture consist of two main sections, feature extraction and classification. The former incorporates a number of filters, each being a tensor of problem-specific dimension withholding real values as their elements, and the latter is essentially a fully-connected feedforward artificial neural network.\\
As for using CNNs for solving problems, once the problem's data is obtained and the task is determined, it is now the job of the CNN to receive any instance of the problem's data and report back a result. For working properly, meaning to report back accurate results, a CNN must undergo a process of training, that is optimizing its filter and connection weight values such that the network yields an accurate result most of the time. Depending on the task, a CNN's training process consists of inputting instances of the problems data into the network and modifying it based on the error of its performance. Hence the term "learning".\\
\\
\\Genetic Algorithm (GA), as one of the subsets of Evolutionary Algorithms, is a global optimization method inspired by the process of natural selection for solving both constrained and unconstrained optimization problems. Genetic algorithm repeatedly modifies a population of individual solutions by selecting the best individuals from the current population as parents and using them to produce children for the next generation through a number of bio-inspired operators. Over successive generations, the population "evolves" towards an optimal solution. As the training process of a CNN is basically an optimization problem, intuition suggests that GA can be used to do that.\\
In the interesting literature of training CNNs with genetic algorithm, to the best of our knowledge, most of the respective works use the same network to chromosome encoding. This encoding proves rather good for smaller sized Multilayer Perceptrons (MLP) and simple Auto-Encoders. But as the networks get larger and the number of their training parameters grow, specially in case of CNNs, this encoding proves to be ineffective, as we will demonstrate. Moreover, vague and obscure description of the parameters involved in the GA used for training the networks, as well as a lack of exploration in the different kinds of GA schemes that can be involved in the training process, are some of the most important shortcomings of the current state of the literature.\\
In this work, we attempt to train two different deep convolutional neural network architectures doing an image classification task over two different modern datasets using methods based on genetic algorithm. These methods are carried out using three individual GA schemes, Steady-State, Generational, and Elitism. Our training methods involve novel genetic operators for crossover and mutation. In addition, we introduce the Accordion chromosome structure, an innovative encoding paradigm of the networks to chromosomes that reduces the chromosome's size by a large factor, leading to faster operations time.\\
To evaluate our work, first, our encoding paradigm was compared to traditional network-to-chromosome encoding on how effective they are when being used in a GA based training of the mentioned CNN architectures. Additionally, this comparison was carried out across the three mentioned GA schemes to determine which one of the schemes performs better in means of convergence rate and overall accuracy. Finally, our proposed encoding used alongside the superior GA training scheme is compared to training the mentioned networks with backpropagation training method using Adam \cite{kingma2014adam} optimizer.\\
To summarize, the contributions of this work are as follows:
\begin{itemize}
    \item Introducing the the Accordion chromosome structure. A network to chromosome encoding paradigm that in comparison with traditional encoding methods, produces a chromosome that is much smaller in size by a large factor, leading to faster operations time and a more effective evolution process.
    \item Proposing novel genetic operators involved in the GA-based training of the networks.
    \item Introducing three innovative genetic algorithm schemes for training deep CNNs and performing a thorough evaluation of their performance against each other and two backpropagation training methods.
\end{itemize}
The remainder of this work is structured as follows. \autoref{sec:background} provides a detailed background on convolutional neural networks and genetic algorithm. A brief summary of other related works in the literature can be found in \autoref{sec:related}. The proposed methodologies are fully described in \autoref{sec:method}. \autoref{sec:result} shows all the results reported from our implementations and comparisons. And finally, the conclusions are brought in \autoref{sec:conclusion}.
\section{Background}\label{sec:background}
\subsection{Convolutional Neural Networks}
\begin{figure}[h]
  \centering
  \includegraphics[width=16.7cm, height=3.7cm]{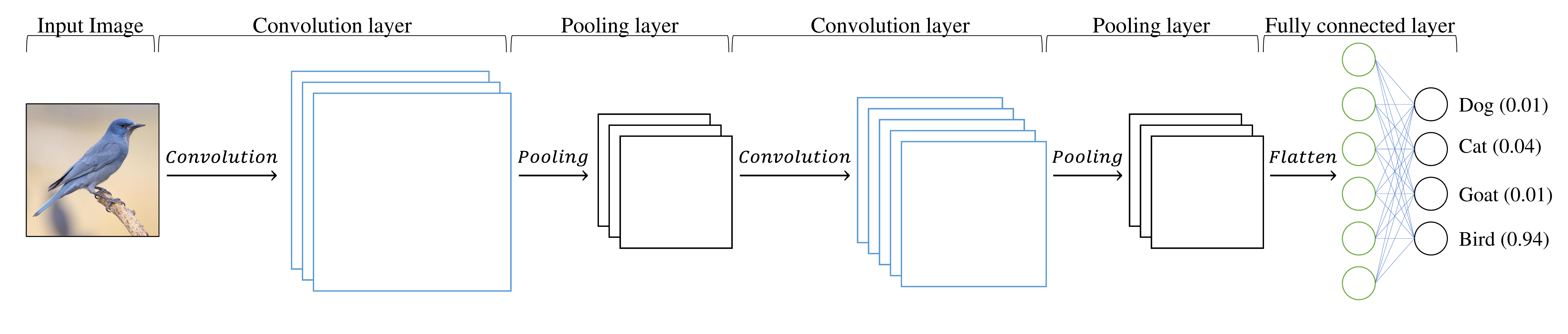}
  \caption{\small A simple convolutional neural network with two convolution layers, two pooling layers, and a single fully connected (dense) layer for the classification of images into four classes; dog, cat, goat, and bird.}
  \label{fig:CNN}
\end{figure}

\begin{figure}[h]
  \centering
  \includegraphics[width=10.2cm, height=5cm]{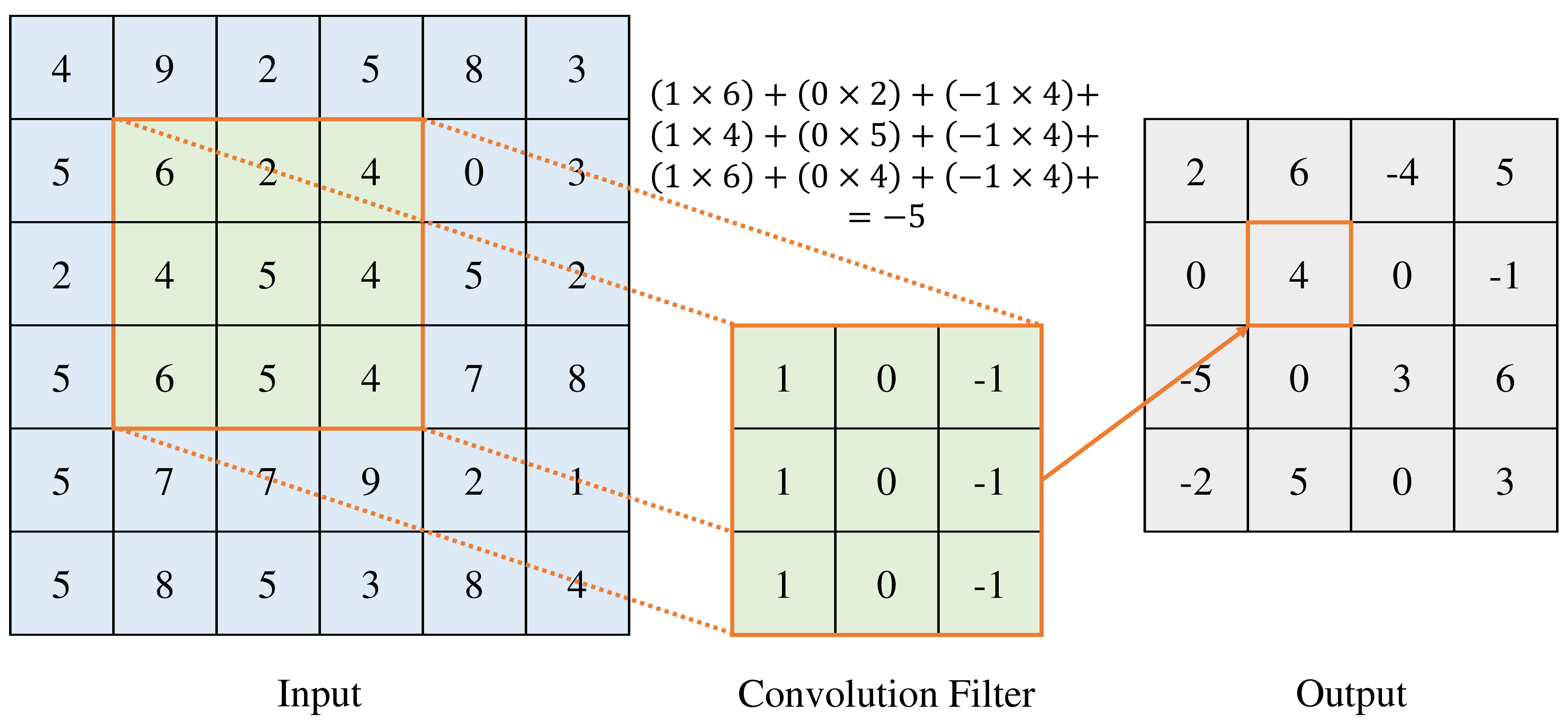}
  \caption{\small A convolution filter being applied to an image.}
  \label{fig:convfilter}
\end{figure}

\begin{figure}[h]
  \centering
  \includegraphics[width=6.2cm, height=4cm]{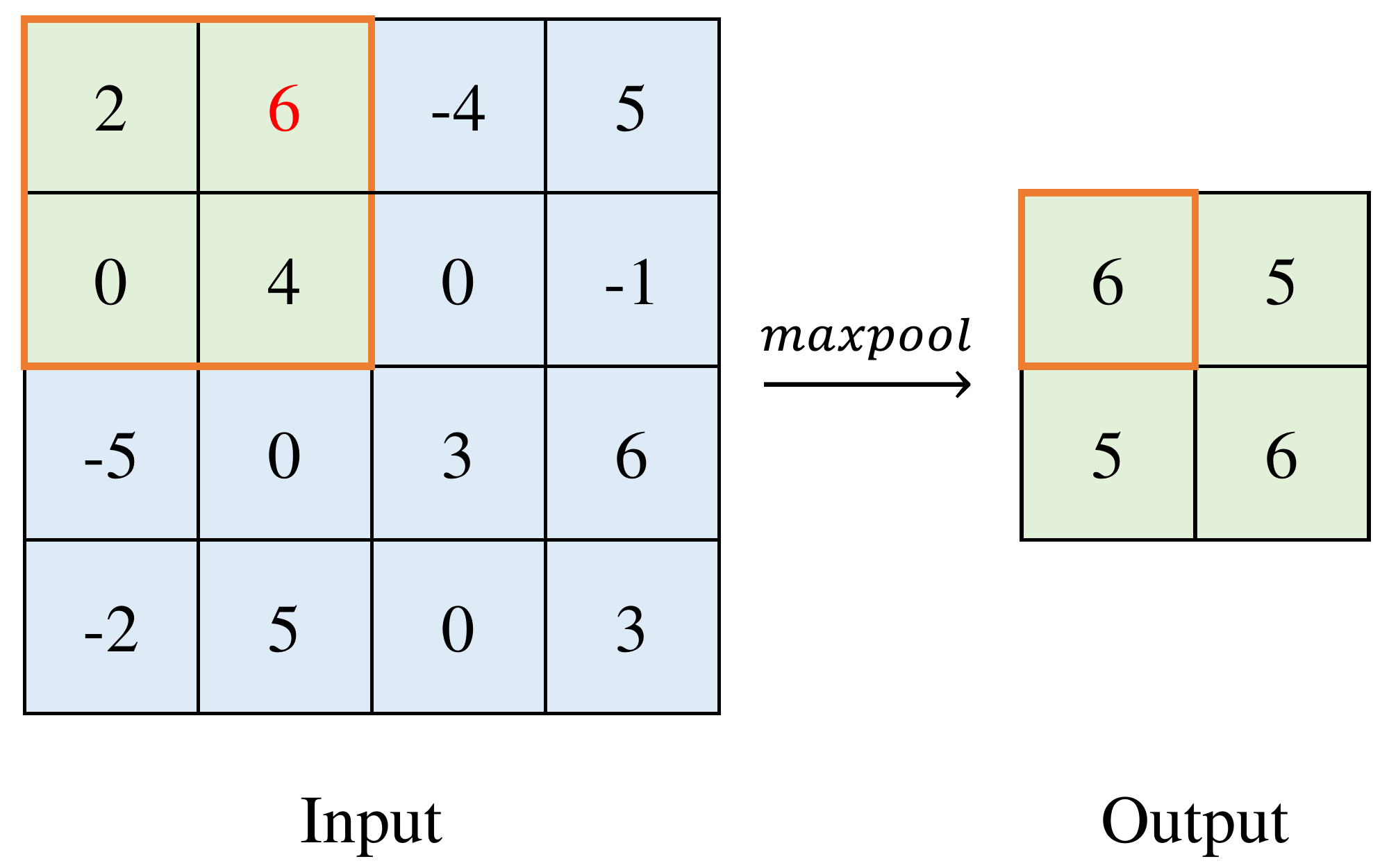}
  \caption{\small A $2\times2$ maxpooling filter, passing the element with the largest value at each of its positions.}
  \label{fig:maxpool}
\end{figure}

\begin{figure}[h]
  \centering
  \includegraphics[width=5.9cm, height=4.2cm]{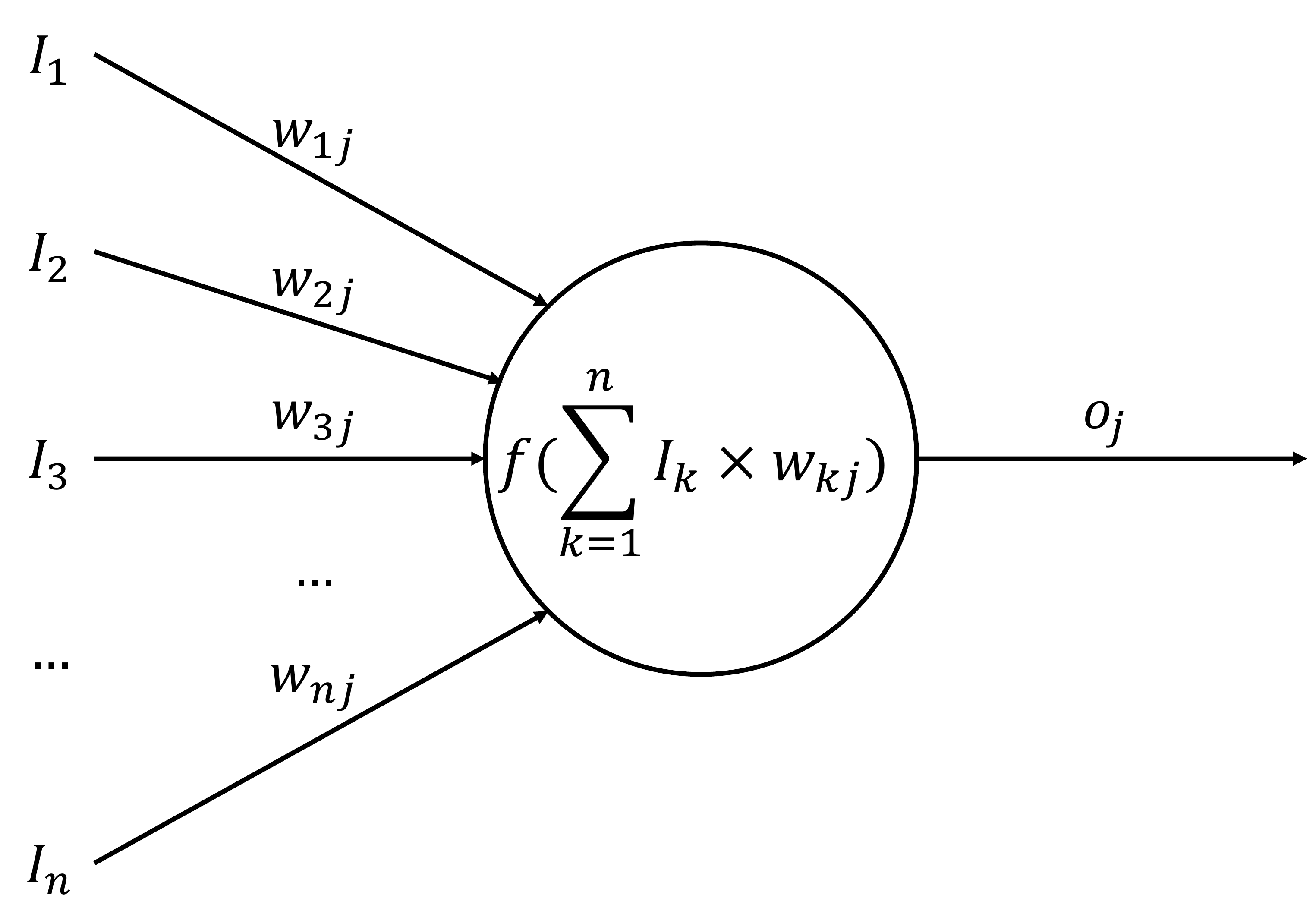}
  \caption{\small An artificial neuron's layout, receiving \textit{n} inputs, calculating their weighted sum, and outputting $o_j$, the transformed weighted sum applied by the transform function $f$.}
  \label{fig:Neuron}
\end{figure}

Convolutional neural networks are a specialized kind of artificial neural networks used widely in the field of image and video analysis. They are excellent tools for finding patterns which are far too complex or numerous for a human programmer to extract and teach the machine to recognize.
Artificial neural networks themselves are biologically inspired computation systems intended to replicate animal brains. They are made up of simple processing elements, called neurons, that interact with each other using a network of weighted connections. Artificial neural networks are a remarkable method for classifying noisy and uncertain data and they can be trained to exceptional accuracies in a very desirable time. However, as the problem gets complicated and the network's input size starts to grow, they fail to scale their performance. This failure arises from their inability to detect certain important features in the data, and instead, trying to report results using only the raw data. So, the model's designer has to realize these important features into the networks manually. But in most cases, these features are way to complex to be coded into a model. Or even worse, they might be overlooked or undetected. To address this issue, convolutional neural networks come to the rescue. An illustration of a simple convolutional neural network is depicted in figure \autoref{fig:CNN}.
\\A convolutional neural network is comprised of two sections, feature extraction and classification. The first section incorporates a number of layers to find and detect certain attributes in an image, such as edges, shapes and \textit{etc.} By detecting these certain attributes in an image, or to put it more precisely, by detecting the existence probability of these certain attributes, the feature extraction section reduces the dimension of the input image into something meaningful and small enough for the classification section to classify. The feature extraction section holds layers such as convolution, Rectified Linear Unit (ReLU) and pooling. In a convolution layer, a filter of predefined sized is applied to its input image, by multiplying the filter values with their corresponding image values and then summing everything up. This sum is then passed through a ReLU function and the resulting value is then placed in the reduced image target pixel. Then, this filter moves through the entire image to extract its important traits. The operation of a convolution filter is shown in \autoref{fig:convfilter}. Next, a pooling operation happens to the resulting image to further reduce its dimension. This operator is also depicted in \autoref{fig:maxpool}. This filtering and pooling processes are then repeated until the input image reduces to acceptable dimensions. Next, it is the job of the classification section to classify this reduced image.
\\The classification section is a simple multilayer perceptron that seamlessly receives the dimensionally reduced image as its inputs and outputs the probabilities of that image belonging to certain classes. In its structure, a multilayer perceptron has a number of hidden layers made up of artificial neurons (or nodes) that interact with each other using weighted connections. These neurons are simple processing elements applying a predefined transform function to the weighted sum of their inputs. As it is depicted in \autoref{fig:Neuron}, a neuron $j$ operates by receiving inputs $I_i$ (which are the outputs of other neurons) through its connections and then calculating their weighted dot product as $\sum_{k=1}^{n}{I_k.w_{kj}}$.\\
Next, the product value is transformed using the transform function $f$ and then broadcasted through all of the neuron's outgoing connections as $o_j$. some of the most popular transform functions are Sigmoid function with the formula $S(x)\ =\ \frac{1}{1+e^{-x}}$ and Rectified Linear Unit (ReLU) defined as $f(x)\ =\ max(0,\ x)$.
\\So, once an instance of the problem's data is inputted through the network, that data instance is flowed and changed through each filter, layer and neuron of the network until reaching the last layer and eventually, being outputted as some values. To measure the performance of the network, a proper loss function must be chosen to evaluate how much the output is correct for any given labeled input instance. By changing the parameters of the network, specially its filter and weight values, the error of performance (\textit{i.e.} the value of loss function) for different instances of the problems data changes. Optimizing this value configuration so the loss function reports back the lowest possible value is known as the training of a convolutional neural network.
\subsection{Genetic Algorithm}
In computer science and operations research, Genetic Algorithm (GA) is a metaheuristic inspired by the process of natural selection that was introduced by John Holland in 1960 based on the concept of Darwin’s theory of evolution. GA is most commonly used to generate high-quality solutions to optimization and search problems by relying on bio-inspired operators such as mutation, crossover and selection. \cite{holland1992adaptation}\\
In a genetic algorithm, a population of solutions (called individuals, members, \textit{etc.}) to an optimization problem is evolved towards better solutions. Each solution, encoded with real or binary numbers or some other encoding, is represented by a chromosomes which can be changed and altered. The evolution, which is an iterative process, usually starts from a population of randomly generated individuals, with the population in each iteration called a generation. In each generation, the fitness measurement of each and every individual in the population is evaluated, which is usually the value of the objective function in the optimization problem being solved. The more fit individuals are selected (with a high chance) from the current population to be modified (recombined and possibly randomly mutated) for the formation of a new generation. The new generation of candidate solutions is then used in the next iteration of the algorithm. Commonly, the algorithm terminates when either a maximum number of generations has passed, or a satisfactory fitness level has been reached for the population.\\
A typical genetic algorithm requires a genetic representation of the solution domain and a fitness function to evaluate the solution domain. A standard representation of each solution is by using an array. The main property that makes this genetic representations convenient is that its parts are easily aligned due to their fixed size, which enables simple crossover operations. Once the genetic representation and the fitness function are defined, the algorithm proceeds to initialize a population of solutions and then improving it through repetitive application of the selection, crossover, and mutation operators.
Next describes the flow of how a genetic algorithm generally operates. This is shown in \autoref{fig:GAFlow}.
\begin{enumerate}
    \item Initialization: The initial population contains several hundreds or thousands of possible solutions often generated randomly. This allows a desirable sampling of the entire range of possible solutions.
    \item Evaluation: A fitness function is defined over the genetic representations of the solutions to assign a fitness value to each member based on measuring the quality of the represented solution according to some performance evaluation criteria. The fitness function is problem dependent, henceforth making its correct determination an important step in the configuration of a working genetic algorithm.
    \item Selection: During each successive generation, whole or a portion of the current population is selected to reproduce a new generation. Solutions involved in the reproduction are selected through a fitness-based process, where fitter solutions are typically more likely to be selected. Some examples of selection strategies are fitness-proportion selection (or most commonly known as roulette-wheel selection), tournament selection, and stochastic universal sampling.
    \item Genetic Operators: The next step is to create a next generation population of solutions from the selected solutions through a combined use of genetic operators such as crossover and mutation. For the production of each new solution, a pair of "parent" solutions (or sometimes, just one solution) is selected from the pool of previously selected fit individuals for breeding. By producing a "child" solution using the mentioned methods of breeding, a new solution is created which mostly shares many of the characteristics of its "parent(s)". Again, new parent(s) are selected for each new child, and this process continues until a new population of solutions of appropriate size is generated. In order to produce a new generation of solutions, this newly bred children can replace some or the entirety of the members of a current generation. These processes ultimately result in the next generation population of chromosomes that is different from the previous generation. Generally, the average fitness of a population increases by this procedure, since only the best organisms, along with a small proportion of lesser fit solutions, are selected for breeding. These less fit solutions ensure genetic diversity within the genetic pool of the parents and therefore, ensure the genetic diversity of the subsequent generation of children. Although crossover and mutation are known as the main genetic operators, it is possible to use other operators such as regrouping, colonization-extinction, or migration in genetic algorithm.
    To find a reasonable configuration of settings for the problem class being worked on, parameters such as the mutation probability, crossover probability and population size must be tuned properly. A very small mutation rate may lead to genetic drift. A recombination rate that is too high may lead to premature convergence of the genetic algorithm. A mutation rate that is too high may lead to loss of good solutions. So, the correct configuration of tuning these parameters is an important step towards making a good genetic algorithm.
    \item Termination: The above process is repeated until some termination condition has been reached. Common terminating conditions are finding a solution that satisfies minimum criteria, reaching a fixed number of generations, or reaching a plateau in the highest ranking solution's fitness.
\begin{figure}[h]
  \centering
  \includegraphics[width=3.9cm, height=7cm]{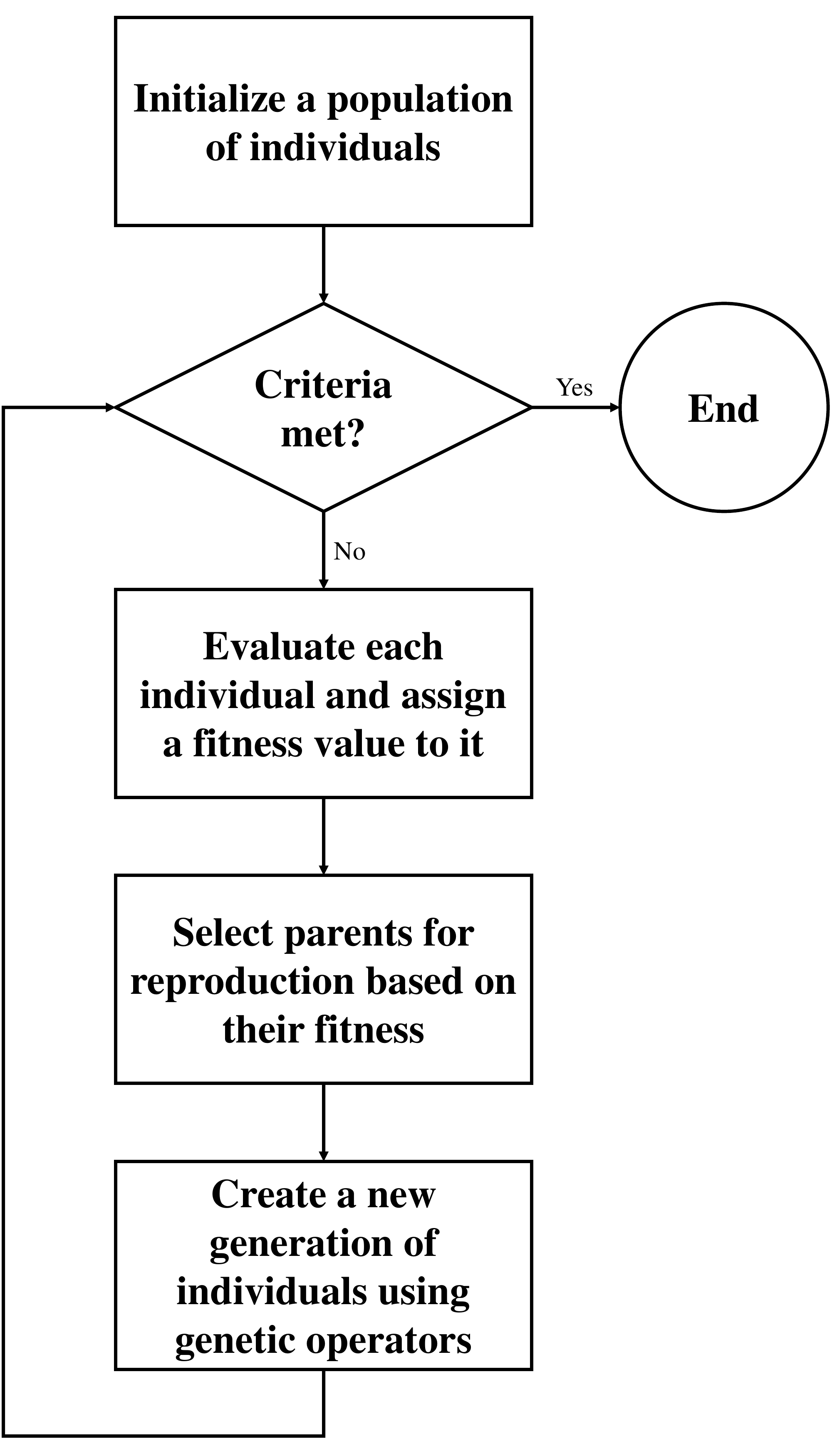}
  \caption{\small The general flowchart of a genetic algorithm.}
  \label{fig:GAFlow}
\end{figure}
\end{enumerate}
\section{Related Work}\label{sec:related}
Various works have been presented in the field of combining genetic algorithm with feed-forward artificial neural networks that target the different aspects of this combination.
For instance, \cite{miller}, \cite{Leung} attempt on employing genetic algorithm for designing and evolving an artificial neural network's structure, meaning its architecture and parameters. Whereas \cite{jaddi}, \cite{kitano}, \cite{koza}, \cite{stanley}, \cite{zanchettin}, \cite{palmes} introduce methods for the concurrent optimization of both an artificial neural network's structure and weights.
Moreover, many have proposed variations of genetic algorithm for the sole purpose of training a neural network (\textit{i.e.} weight optimization) while addressing both the genetic algorithm and backpropagation's advantages and disadvantages. For example,~\cite{montana,montana2}, \cite{david}, \cite{gupta,sexton}, \cite{kattan} report the superiority of genetic algorithm, while on the other hand, \cite{che} reports in favor of backpropagation.\\
\\
Several works have propose the involvement of genetic algorithm in the optimization of convolutional neural network's structure and weights. \cite{xie} has introduced a genetic algorithm for learning deep CNN's structures, as well as a fixed-length binary encoding of the network's structures. For evolving the architecture and the connection weight values initialization of deep CNNs, \cite{sun} has proposed a new method using genetic algorithm and a variable-length encoding paradigm of the networks. Likewise, \cite{young} introduced MENNDL, an evolutionary algorithm based framework for optimizing the hyper-parameters of a deep CNNs.\\
Focusing on using genetic algorithm to train CNNs, \cite{oullette} proposed a genetic algorithm of a three-layer deep CNN used for detecting cracks in images of different structures. Furthermore, \cite{kitano}, \cite{cheung}, \cite{kanada}, \cite{alba}, \cite{mcinerney}, \cite{yin} have presented methods of using evolutionary methods, including genetic algorithm, for the training feedforward artificial neural networks and convolutional neural networks.\\
While all of the respective mentioned works deliver an extraordinary contribution to the field, they all fall short in some very important aspects. The lack of a thorough comparison of their method's performance on different modern deep networks and datasets and leaving out the exploration of different variations of GA parameters are some of the shortcomings of these works. Most importantly, the method that most of these works propose to bring a network's structure into the domain of genetic algorithms (\textit{i.e.} encoding to chromosome), while helpful and inspiring, proves to be ineffective, since it becomes to large as the network's size and parameters grow. That's why in this work, we make an effort to construct means of addressing these shortcomings, by proposing a new encoding paradigm that shortens the resulting chromosome by a large factor, leading to faster operation times, and additionally, introducing GA based methods carried out in three different schemes for training deep CNNs that are obtained through a comprehensive assessment of alternating the combination of the GA parameters involved in the training process.\\
To summarize, our work brings improvement to the literature, by proposing an encoding paradigm that handles larger networks better and scales well as they grow. And furthermore, employing this encoding alongside novel genetic operators for a GA based training of deep CNNs carried out using three individual schemes. In the next sections, we will describe our work and obtained results in full detail.
\section{Methodology}\label{sec:method}
\subsection{Genetic Algorithm for Training Convolutional Neural Networks}
\begin{figure}[h]
  \centering
  \includegraphics[width=16cm, height=10cm]{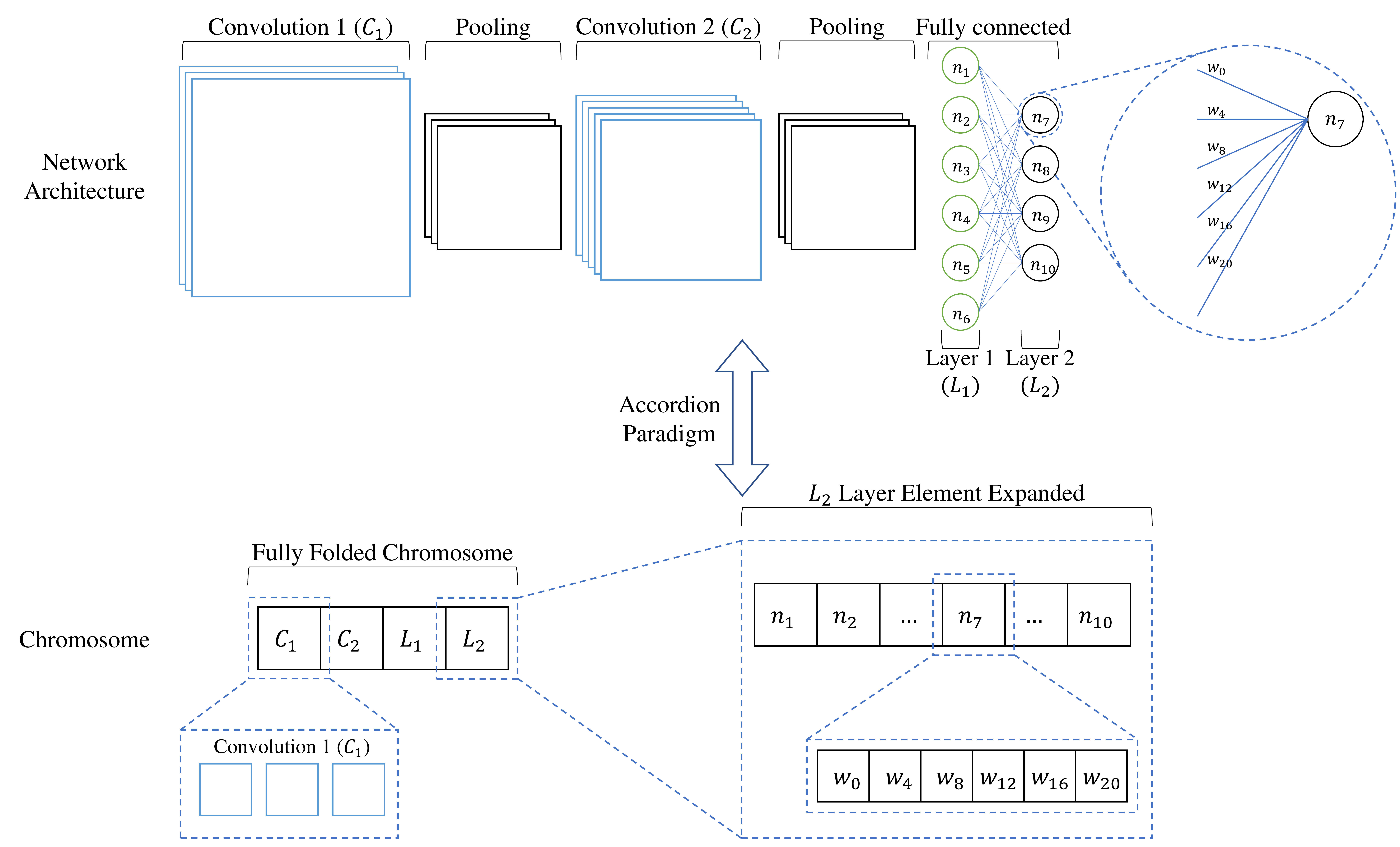}
  \caption{\small Our proposed encoding paradigm of a network into a chromosome.}
  \label{fig:encoding}
\end{figure}

The challenge of using genetic algorithm for training a deep convolutional neural network is in mapping the problem from the domain of artificial neural networks literature to the domain of genetic algorithm literature. Meaning, how GA can be used to train a network. The initial intuition suggest that the networks should act as members of a population for our algorithm. A network's performance, somehow, should represent the network's fitness, and the algorithm's iterative process should "evolve" the population of networks towards better accuracies.\\
The first part of the challenge is the encoding problem, \textit{i.e.} how to map a network to a chromosome. Traditional encoding paradigms extracts every single trainable parameter in a network's structure and then encapsulate them as an array construct to achieve a chromosome representation of the network. In this manner, each element of the chromosome array holds only a single network parameter. For example, encoding the simplest MLP for the MNSIT classification task (2 hidden layers each holding 512 neurons, $28\times28=784$ input neurons and 10 output neurons, leading to a total of 668672 trainable parameters) leads to chromosome with the size of 668672 elements, each holding one of the trainable parameters. As long as a network is small, this representation works quite well. But as we know, today's networks are not this small in size. Modern CNNs can hold up to millions of trainable parameters. A traditional encoding of these large networks leads to an ineffective chromosome structure, that is to large to handle. Furthermore, a smaller chromosome size benefits from faster, and hopefully, more impactful genetic operators.
To handle large sized networks, an encoding is required that can somehow squeeze the large number of trainable parameters into a construct with suitable size. Inspired by the work of \cite{montana}, this can happen by considering the representation that is an array consisting the network's convolution filters and fully connected layers as a whole in its elements. Meaning, each element of the chromosome holds either a layer full of its neuron's ingoing connection weight value or the entirety of a convolution filter with all of its values. This new encoding paradigm provides a promising decrease in the chromosome representation size of the networks that leads to faster operation times, specially as the networks grow larger in size. As we demonstrate in our results, the use of this encoding leads to a faster convergence rate of the evolution process and an eventual higher accuracy threshold. This encoding paradigm is illustrated in \autoref{fig:encoding}.
\\Now that the chromosomes are considered, a fitness value for its representing network's performance must be assigned to each chromosome. This value can be the network's accuracy, Mean-Squared-error (MSE), Sum-Squared-error (SSE), \textit{etc}. Here, we used categorical cross-entropy. Next, an appropriate number of chromosomes would gather to create a population of solutions, each one representing a network similar with others in only structure parameters, but different in convolution filter, connection weight values, and performance measurement. Now, a genetic algorithm can be used to train this population. A very general approach to this endeavor is as:
\begin{enumerate}
    \item Initializing a population of individual chromosomes, each representing a network, using arrays with appropriate length which some of its elements hold tensors of predefined size filled with random values as the network's convolution filters and the rest of the elements holding the fully connected section layers, each of them expanding into randomly chosen values for the ingoing connection weight values for each neuron in that layer.
    \item Assigning a fitness value to each member of the population based on the evaluation of it's representing network.
    \item Selecting individuals based on their fitness value as parents for reproduction of a new generation of individuals.
    \item Reproduction of new children by conducting crossover and mutation operations between selected parents.
    \item Repeating from step 2 until satisfaction.
\end{enumerate}
With the described methodology, a genetic algorithm can be carried out in three schemes. A Steady-State scheme genetic algorithm develops in a manner which with each iteration, a number of new children (mostly one or two) are produced to replace the worst members of the generation. A Generational scheme genetic algorithm develops to replace the whole population with new children at each iteration. But an Elitism genetic algorithm scheme creates the next generation of population by keeping some of the best and fittest members of the current generation (the elites) and passing it on to the next generation, and creating the remainder of the needed population through a breeding process between the generation's fitness-based selected members. These schemes will now be described in full detail.
\subsubsection{Steady-State Genetic Algorithm for training Convolutional Neural Networks}\label{sec:steadyscheme}
The steady-state scheme for training a convolutional neural network consists of the following steps:
\begin{figure}[h]
  \centering
  \includegraphics[width=9cm, height=9cm]{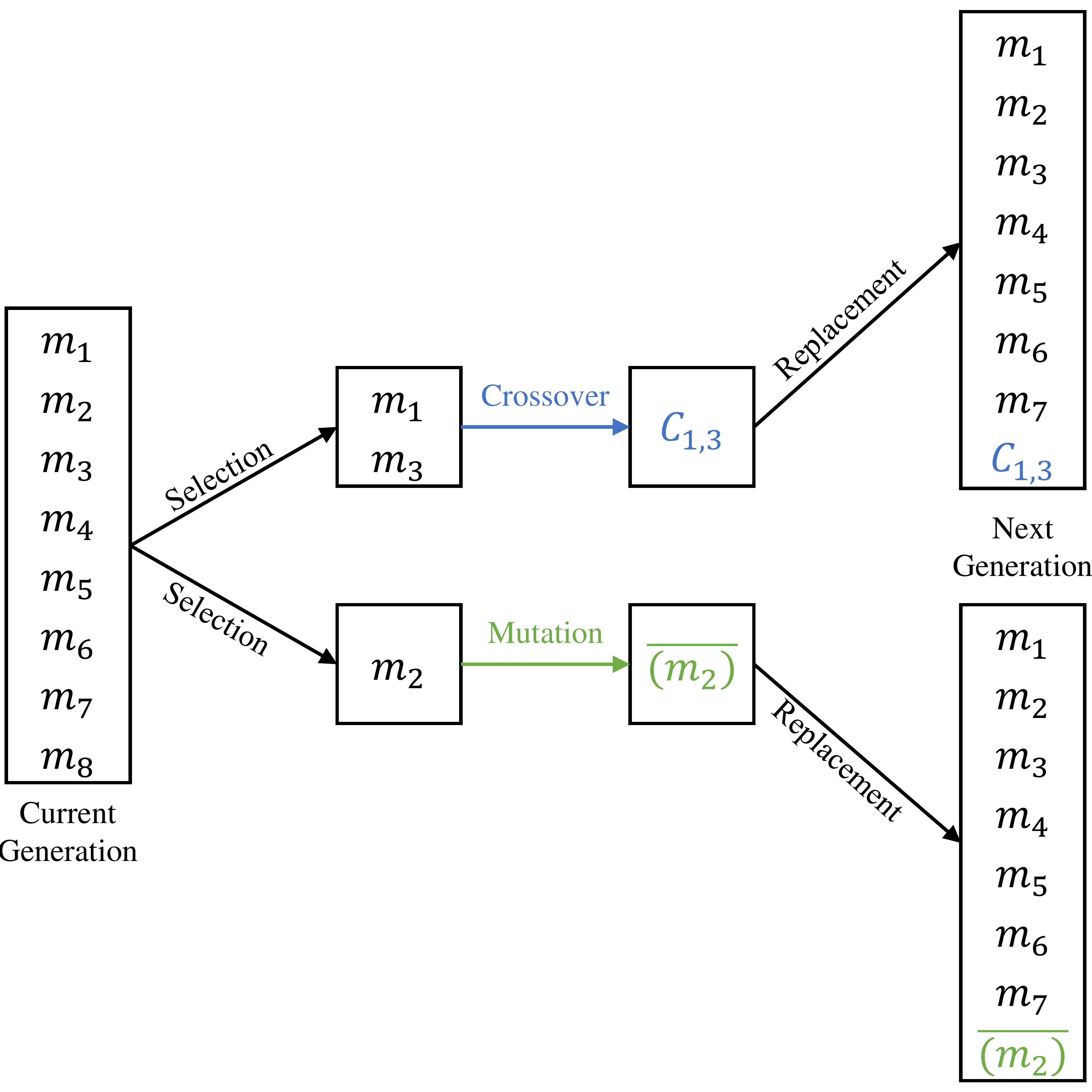}
  \caption{\small The steady-state scheme genetic algorithm for training a convolutional neural network.}
  \label{fig:steady}
\end{figure}
\begin{enumerate}
\item \textit{Initialization:} In this step, some networks equal to \textit{pop\_size} are initialized using Keras, with their convolution filter and connection weight values being assigned with a random number drawn from a truncated normal distribution centered on zero using the Keras' built in Glorot normal initializer.
\item \textit{Evaluation:} In the duration of this step, each network's performance is evaluated based on its accuracy reported by Keras' \textit{model.evaluate()} function. This particular step is utilized in a parallel fashion using the \textit{multiprocess} library in Python, allowing for a faster program run time.
\item \textit{Fitness Assignment:} Each network is assigned a fitness value $f_{i}$ based on its evaluation. Here, we used the accuracy of each network as its fitness value.
\item \textit{Selection:} A selection probability is assigned to each network in this step. For our work, we used one of the most famous selection strategies; Fitness Proportional Selection. Or most commonly known as, Roulette Wheel selection. In this selection, the higher the fitness of a network, the more probable it is to be selected as parent for reproduction. Thus, the probability of each network being selected as a parent for reproduction is,
\begin{equation}
    P_i=\frac{f_i}{\sum_{i=0}^{pop\_size}f_i}.
\end{equation}
We can easily deduce the fact that $\sum_{i=0}^{pop\_size}P_i=1$.\\
Now, the \autoref{tab:table} can be constructed, which holds the index number, fitness value, and selection probability value for each member of the population in a sorted fashion during each generation. This table is used numerous times during the program's run to store a network's fitness, calculate its selection probability, and add and/or remove members from the population during the iterative run. 
For the breeding process in this scheme, during each generation, one child is spawned and it is replaced with the least fit member of that generation, thus discarding it.
\begin{table}[h]
 \caption{Table of parameters associated with each network during a generation}
  \centering
  \small
  \begin{tabular}{c|c|c|c}
    Index & Evaluation & Fitness & Selection Probability \\ \hline
    $network_1$        & $eval_1$             & $f_1$           & $P_1$                         \\
    $network_2$       & $eval_2$              & $f_2$           & $P_2$                         \\
    ...       & ...              & ...           & ...                         \\
    $network_{pop\_size}$         & $eval_{pop\_size}$              & $f_{pop\_size}$           & $P_{pop\_size}$ 
  \end{tabular}
  \label{tab:table}
\end{table}
\\Our crossover operator produces one child from two parents and the mutation operator creates a single child from one parent. So the algorithm needs to decide on how it is going to spawn a new member; using crossover or mutation. A fair or unfair coin-flip decision making can be used here. If crossover is selected, the selection process will select two members as parents of a new child and if mutation is selected, the process selects one member that undergoes mutation. If mutation is selected, The selection process traverses the selection probability column. Then it generates a random number drawn from the uniform random distribution in the interval [0, 1] and compares it to \(P_i\). If the random number is smaller than\ \(P_i\), then the \(i^{th}\) member will be selected for reproduction. This process happens twice if crossover is selected, henceforth, selecting two parents. 
\item \textit{Crossover:} In this process, two parents will spawn a new child sharing some of their attributes. For this operator, the fully folded chromosome is not considered. Instead, a semi-folded chromosome structure, that each of its elements are either the convolution filters or the entirety of the ingoing weights of  neuron in the fully connected section. To simplify, the semi-folded chromosome structure is obtained by expanding the layer elements in the fully folded chromosome. As for the operator, for each convolution filter in the child's network, the process randomly selects one of the parents and then copies the corresponding convolution filter from the selected parent into the child. And for each neuron in the classification section of the child's network, again, the process randomly selects one of the parents and then copies all of the ingoing weights of the equivalent neuron in the parent into the ingoing weights of the destined neuron in the child. This operator description is inspired by the work of \cite{montana}.
 \begin{figure}[h]
  \centering
  \includegraphics[width=16cm, height=2cm]{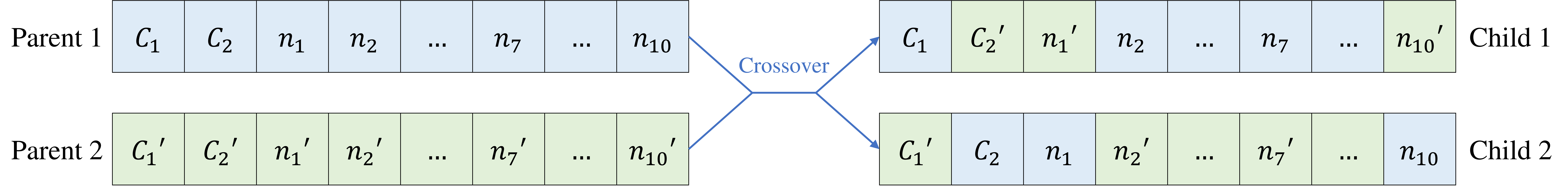}
  \caption{\small A crossover process, producing two children that complement each other. In case of the steady-state scheme, only one of the children is considers. But for the generational scheme, all two are considered.}
  \label{fig:Crossover}
\end{figure}
\item \textit{Mutation:} During this process, one parents will spawn a new child sharing most of its attributes. The process clones an exact copy of the parent's chromosome and randomly selects some of the elements in the semi-folded chromosome structure. If the selected element holds a convolution filter, then for each value in the filter, it is replaced by a Gaussian noise centered around the value with the a derivation of 0.5, and if it is a neuron, it adds to its ingoing weights a random number drawn from the initialization distribution, leading to creation of one new child. This operator description is also inspired by \cite{montana}.
\begin{figure}[h]
  \centering
  \includegraphics[width=16cm, height=0.9cm]{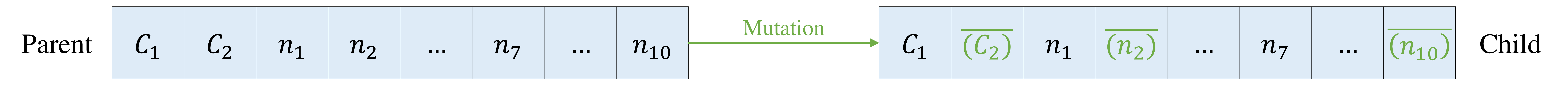}
  \caption{\small A mutation process, spawning one child.}
  \label{fig:mutation}
\end{figure}
\item \textit{Replacement:} The new child is inserted into the next generation of population to replace the least fit member. By this manner, it will discard the replacing member from the population. in this way, \textit{pop\_size} is kept the same throughout the generations. This step marks the completion of one iteration of the steady-state scheme.
\end{enumerate}
The flow above is repeated from step 2 until some criteria (\textit{e.g.} number of generations, accuracy, \textit{etc}.) is met.\\
We can see that this scheme has a benefit of keeping the best solutions intact throughout the generation and only changing the population once a better solution is found. This allows for a stable but slow convergence to the global optimum. An example illustration of this approach for an 8 member population is shown in \autoref{fig:steady}.\\
We used the described steady-state genetic algorithm to train two different architectures of convolutional neural network (\textit{i.e.} to optimize their filter and weight values). It is important to mention that all of the values have an encoding of float32 and no biases and no dropouts were considered in the network for the sake of simplicity.\\
As for the first step of the scheme, \textit{pop\_size} number of networks are created using Keras directives. All of these networks have the same architecture and structure parameters. Meaning all of their convolution filters, pooling, and fully connected sections have the same dimension. But, all of the values in their filters and connection weights are assigned randomly and different. This creates the initial population for our steady-state training method.\\
Now that the initial population is created, each network is evaluated against the problem's dataset, and based on the accuracy that it reports, is assigned its fitness value. Next, the selection probabilities for the roulette wheel selection are calculated and assigned. At each iteration of the algorithm, based on these probabilities, by using a coin-flip decision making, either one parent is selected to produce one off-spring using mutation, or two are selected to produce one off-spring using crossover. The fact is that the networks with better fitnesses might be selected more often.\\
If one parent is selected to have a child using mutation, first, the algorithm creates a network of the same structure but with empty values in their filters and connection weights. Next, it copies the entire parent values into the child, creating an exact clone of the parent. Next, some of the filters and neurons will be mutated according to step 6 of the scheme.\\
If two parent are selected to have one child using crossover, again, the algorithm creates a network of the same structure but with empty values in their filters and connection weights. For each element in the child's chromosome, weather it should hold a filter or a neuron, a coin-flip decision making will happen to determine which one of its parents will be selected to copy that corresponding filter or neuron. Using this crossover, a child will share some of its attributes with one parent and the rest with the other parent. The mentioned process will now be iterated. A population of networks will exist that at each iteration of the algorithm, a new member is added to them. The iteration carries out so the population "evolves" towards better solutions.
\subsubsection{Generational Genetic Algorithm for training Convolutional Neural Networks}
The generational genetic algorithm scheme for training a convolutional neural networks flows as described below. We must note that the Initialization, Evaluation, and Fitness Assignment steps share the exact same traits as the steady-state scheme. So this scheme is described from the step 4.
 \begin{figure}[h]
  \centering
  \includegraphics[width=12cm, height=5cm]{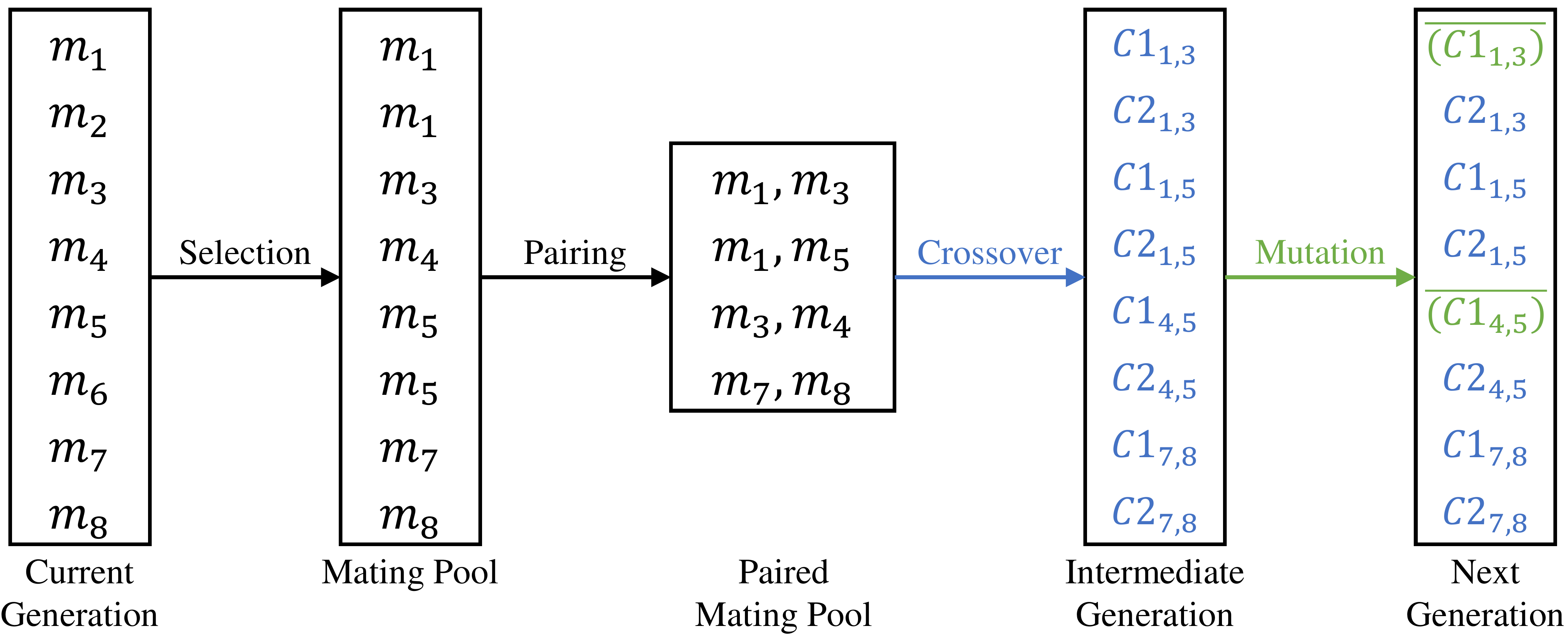}
  \caption{\small The generational scheme genetic algorithm for training a convolutional neural network.}
  \label{fig:generational}
\end{figure}
\begin{enumerate}
\setcounter{enumi}{3}
\item \textit{Selection:} After the fitness value of each member of the population has been determined, a selection probability value must be assigned to each member in order to select suitable parents for breeding. This value is assigned using the same method in the steady-state scheme. Now, some members of the population will be selected for insertion into a mating pool to produce off-springs. There are many different parameters and configurations for a generational scheme genetic algorithm. But for this work, the mating pool holds the same number of members as the population. This means that some members might be selected more than once. Next move involves pairing members in the mating pool, as parents for a children. This pairing is done randomly.
\item \textit{Crossover:} After members of the mating pool are paired, each pair spawns two children. One child is created using the described crossover in the steady-state scheme. The other child is created by complementing the first child. Meaning that for every element in the second child's chromosome, whichever parent was selected for copying its traits from the corresponding chromosome element into the first child, the other parent will now be used to copy its corresponding trait into the second child's element. For example, if the first child has its first chromosome element copied from its second parent, the second child will have that exact element position filled with a copy from the first parent. Here that the mating pool holds the same size as the population, and each pair of parents spawning two children means that now a number of children equal to \textit{pop\_size} are created, ready to be considered as the next generation.
\item \textit{Mutation:} Now that a population of new children are born, some of them will be selected randomly  according to the $mutation\_ratio$ to undergo mutation. The mutation process is the same as the steady-state approach. Except here, the mutated child is not considered a new member, but it is replacing its unmutated self. To put it more simply, the mutation operator for generational scheme is in-place.
\item \textit{Replacement:} The new generation of members completely replaces the last one. This step marks the completion of one iteration of the generational scheme.
\end{enumerate} 
The described flow is repeated from step 2 until meeting some stopping criteria. An example illustration of this approach for a population size of 8 is shown in \autoref{fig:generational}.\\
As for this scheme, the initial population is created, evaluated, and assigned fitness values and selection probabilities same as the previous scheme.
At the next step, based on the selection probabilities, the mating pool is filled with \textit{pop\_size} number of networks. We must remember that networks reporting better accuracies (meaning having better fitnesses) might be selected more than once for entry into the mating pool. As the mating pool is filled, its members are paired randomly. Thus creating $\frac{pop\_size}{2}$ number of pair of parents.\\
The generational genetic algorithm scheme was also used to train different architectures of convolutional neural network. Also here, all of the values have an encoding of float32 and no biases were considered.\\
Once the pairs of parents are realized, each pair will create two children using crossover. This means that for one child, its filters and neuron's ingoing weights will be choosed randomly to be copied from either parent. For each choice, the opposite will be made to construct the second child. Hence in this manner, a pair of parents create two children that complement each other. Each children, being a network itself, is similar in architecture with its parents, but different in the filter and connection weight values. Since there were $\frac{pop\_size}{2}$ number of pairs of parents and each created two children, we now have a number of $pop\_size$ newly bred children.\\
Now some of these children, based on the $mutation\_ratio$, will be selected to undergo the mutation process. In this procedure, each child selected for mutation will have some of its filters and/or neurons mutated according to the mutation operator described in this scheme.\\
By iterating the mentioned process, a population of networks will have their filters and connection weight values changed and altered, in order to "evolve" the entire population towards better accuracies.
\subsubsection{Elitism Genetic Algorithm for training Convolutional Neural Networks}
The elitism genetic algorithm scheme for training a convolutional neural networks carries the following description. Again, the Initialization, Evaluation, and Fitness Assignment steps for this scheme is the same as the steady-state and generational schemes. So, this scheme is also described from step 4.
 \begin{figure}[h]
  \centering
  \includegraphics[width=12cm, height=5cm]{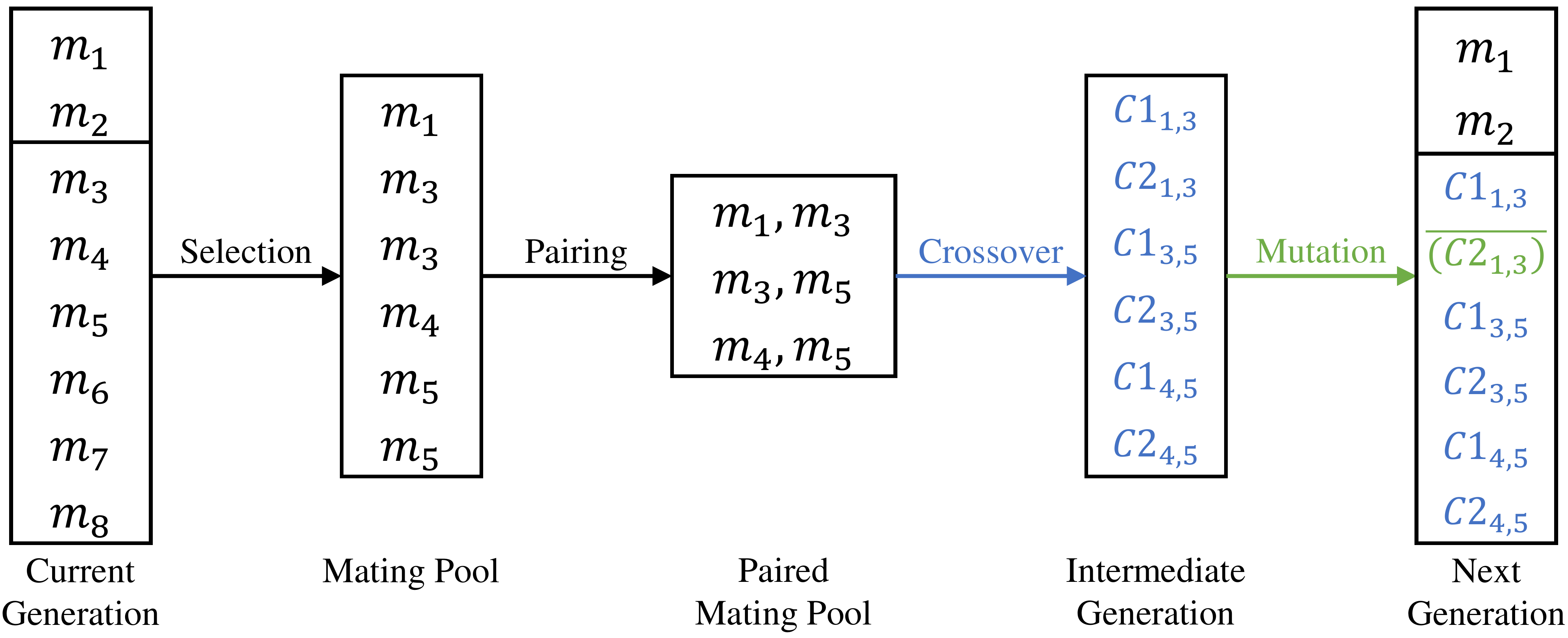}
  \caption{\small The elitism scheme genetic algorithm for training a convolutional neural network.}
  \label{fig:elitism}
\end{figure}
\begin{enumerate}
\setcounter{enumi}{3}
\item \textit{Selection:} Again like the previous schemes, members of the population are assigned with selection probabilities based on their fitness. And again, some members will be selected for insertion into a mating pool to produce off-springs. But unlike the generational scheme, the mating pool does not hold he same number of members as the population. Instead, it holds a portion of it. That's because in this scheme, the fittest members of the current generation (\textit{e.g} top 10) are directly inserted into the next generation. Based on the number of elite members selected, the mating pool will now receive enough members to fill the next generation's population. Unlike the generational scheme, parents in the mating pool will not be paired randomly. instead, for every child that it takes to fill the next generation of population (alongside the previously inserted elites), the algorithm chooses two parents randomly to create ONE child with crossover. After the child is created in the next step, the parents are returned to the pool. This enables some parents to be probably selected more than once to create a child.
\item \textit{Crossover:} Two parents create only one child using the same crossover operator in the steady-state scheme. This child-creating operator will happen until enough new children are born to fill the next generation of population.
\item \textit{Mutation:} Some of the new children created will receive an in-place mutation that is exactly the same as the mutation operator described in the generational scheme.
\item \textit{Replacement:} The new members, accompanying the individually selected and brought in elite members, create the next generation of population. This step marks the completion of one iteration of the elitism scheme.
\end{enumerate} 
This flow is repeated from step 2 until achieving a satisfying stop condition. An example illustration of this approach is shown in figure \ref{fig:elitism} for again, an example small population of size 8. Alongside the previous schemes, the elitism genetic algorithm scheme was also used to train different architectures of convolutional neural network. Once again, all of the values have an encoding of float32 and no biases were considered. And also, the initial population creation, evaluation, and fitness values and selection probabilities assignment are same as the generational scheme.\\
At next, a predefined number of elites are selected to be directly inserted into the next generation of population. For our work, the top 10 fittest members sufficed. Like before, the mating pool is filled with enough members based on their selection probabilities. Again some networks might be selected more than once. After that, based on the number of children required to fill the next generation of population, two parents are selected in random to produce a child using the crossover operator. After that, the parents are returned back into the pool. This means that they might be selected again. Some of these new children will receive a mutation. After that, they are inserted into the next generation. This flow is repeated until the population "evolves" in the better direction.\\
One can see that this scheme possibly provides the best of both previous schemes. On one hand, steady-state scheme is stable but slow. While on the other, the generational scheme "evolves" faster, but it has a more oscillated journey towards optimum solutions. It is hoped here that by combining these two schemes, the best of both can be achieved.
\section{Evaluation Results}\label{sec:result}
\begin{figure}[h]
    \centering
    \begin{subfigure}[b]{0.49\textwidth}
        \includegraphics[width=\textwidth]{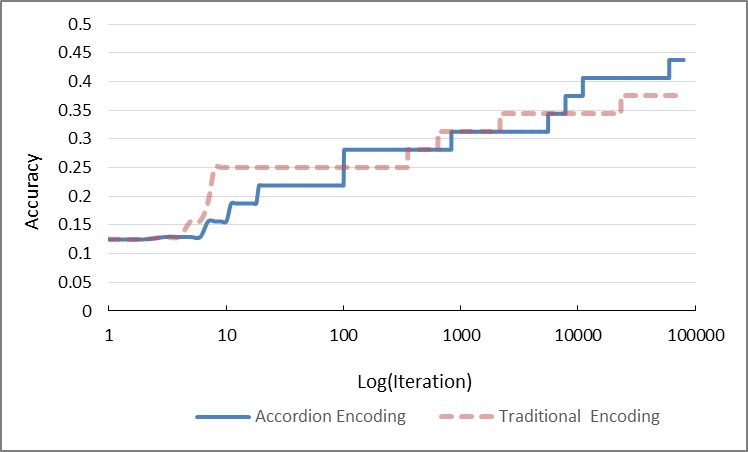}
        \caption{}
        \label{fig:SteadyMNISTEncodingVS}
    \end{subfigure}
    \begin{subfigure}[b]{0.49\textwidth}
        \includegraphics[width=\textwidth]{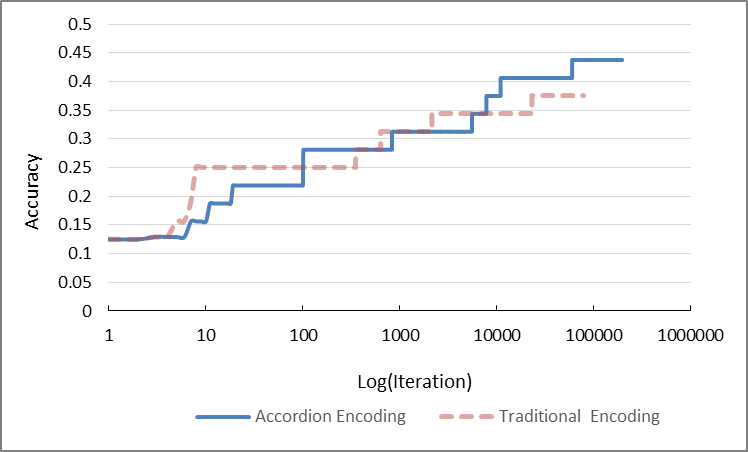}
        \caption{}
        \label{fig:StadyCIFAREncodingVS}
    \end{subfigure}
    \caption{Comparison of the Accordion encoding and traditional encoding used in the steady-state training scheme of (a) the network for the MNIST classification task and (b) the network for the CIFAR10 classification task.}\label{steady}
\end{figure}

\begin{figure}[h]
    \centering
    \begin{subfigure}[b]{0.49\textwidth}
        \includegraphics[width=\textwidth]{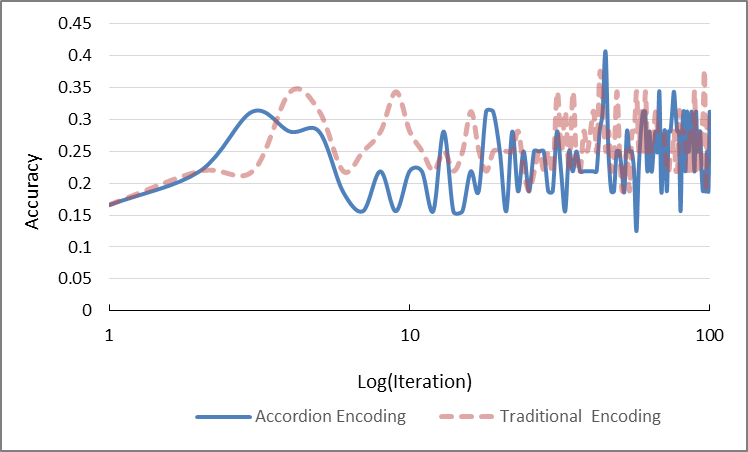}
        \caption{}
        \label{fig:GenerationalMNISTEncodingVS}
    \end{subfigure}
    \begin{subfigure}[b]{0.49\textwidth}
        \includegraphics[width=\textwidth]{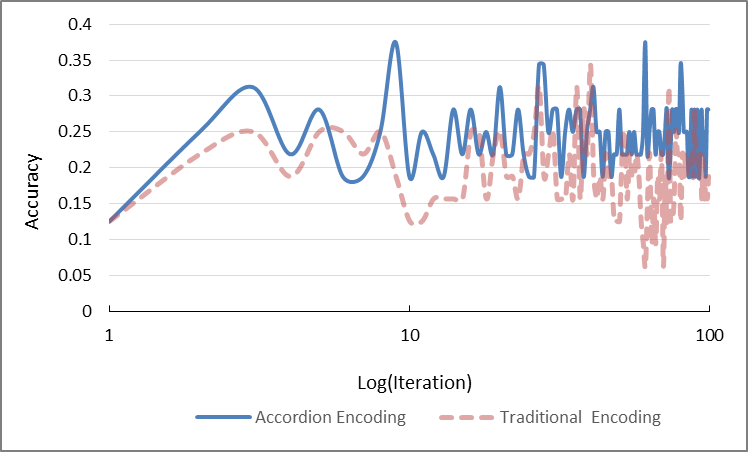}
        \caption{}
        \label{fig:GenerationalCIFAREncodingVS}
    \end{subfigure}
    \caption{Comparison of the Accordion encoding and traditional encoding used in the generational training scheme of (a) the network for the MNIST classification task and (b) the network for the CIFAR10 classification task.}\label{steady}
\end{figure}

\begin{figure}[h]
    \centering
    \begin{subfigure}[b]{0.49\textwidth}
        \includegraphics[width=\textwidth]{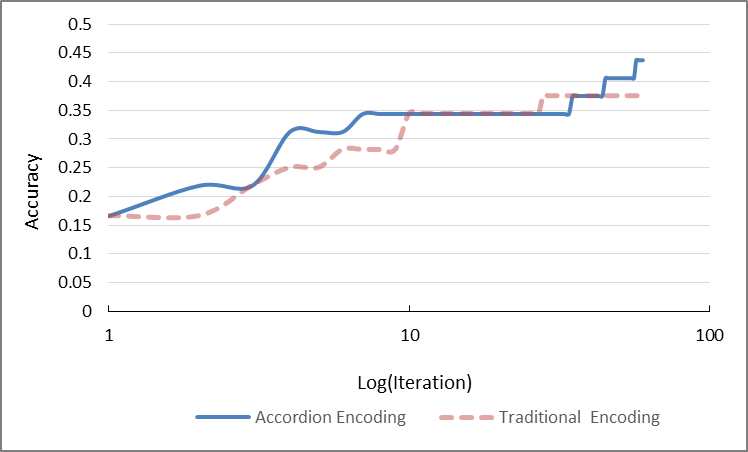}
        \caption{}
        \label{fig:ElitismMNISTEncodingVS}
    \end{subfigure}
    \begin{subfigure}[b]{0.49\textwidth}
        \includegraphics[width=\textwidth]{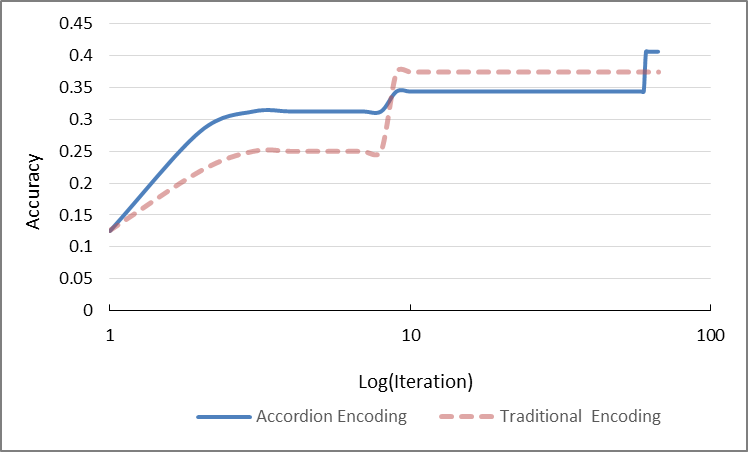}
        \caption{}
        \label{fig:ElitismCIFAREncodingVS}
    \end{subfigure}
    \caption{Comparison of the Accordion encoding and traditional encoding used in the elitism training scheme of (a) the network for the MNIST classification task and (b) the network for the CIFAR10 classification task.}\label{steady}
\end{figure}

\begin{figure}[h]
    \centering
    \begin{subfigure}[b]{0.49\textwidth}
        \includegraphics[width=\textwidth]{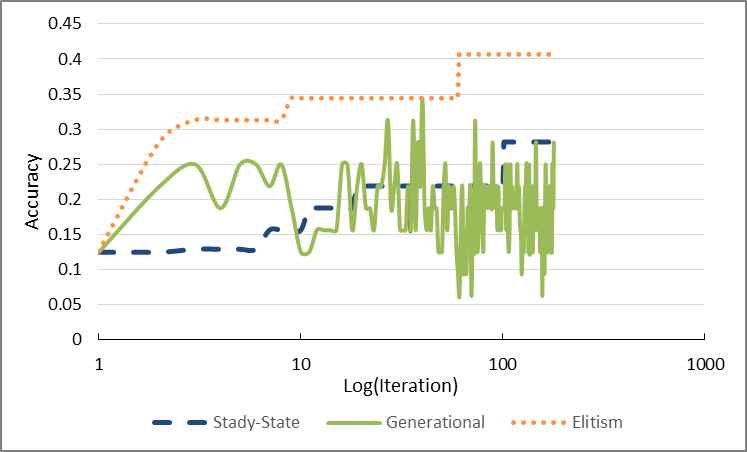}
        \caption{}
        \label{fig:ComulativeMNIST}
    \end{subfigure}
    \begin{subfigure}[b]{0.49\textwidth}
        \includegraphics[width=\textwidth]{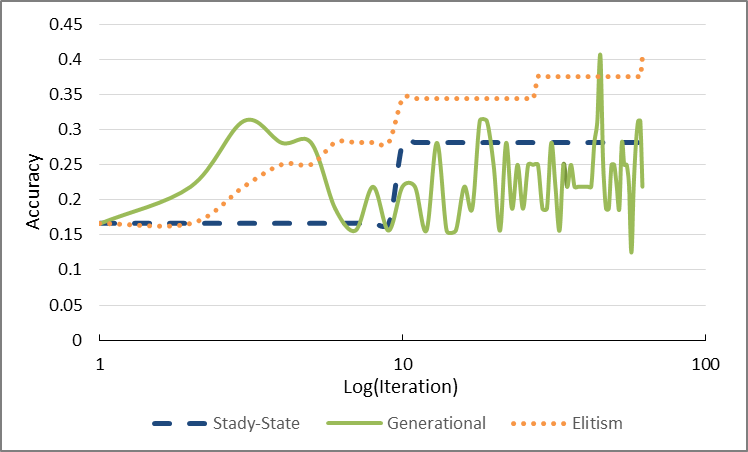}
        \caption{}
        \label{fig:ComulativeCIFAR}
    \end{subfigure}
    \caption{Comparison of the Accordion encoding and traditional encoding used in the elitism training scheme of (a) the network for the MNIST classification task and (b) the network for the CIFAR10 classification task.}\label{steady}
\end{figure}

\begin{figure}[h]
    \centering
    \begin{subfigure}[b]{0.49\textwidth}
        \includegraphics[width=\textwidth]{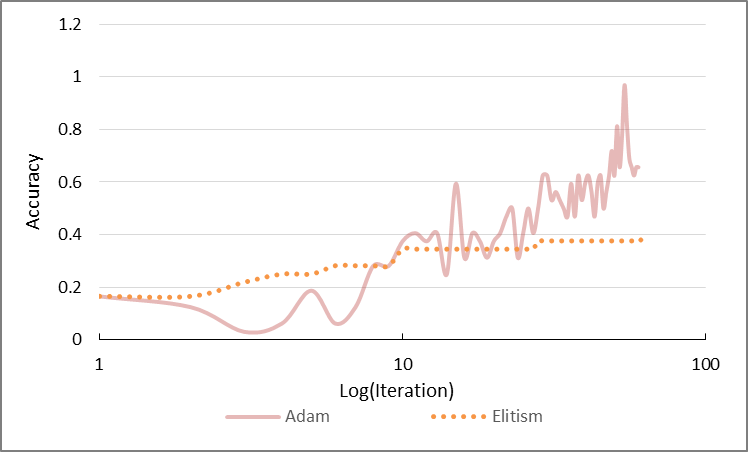}
        \caption{}
        \label{fig:MNIST}
    \end{subfigure}
    \begin{subfigure}[b]{0.49\textwidth}
        \includegraphics[width=\textwidth]{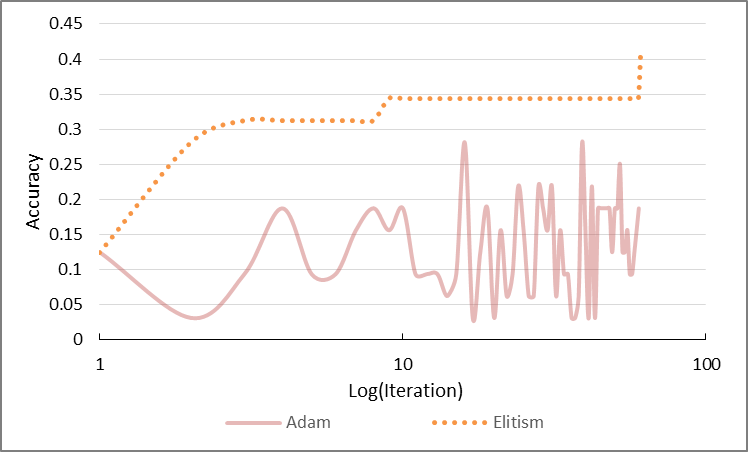}
        \caption{}
        \label{fig:CIFAR}
    \end{subfigure}
    \caption{Comparison of the Accordion encoding used alongside the elitism training scheme and Adam training of (a) the network for the MNIST classification task and (b) the network for the CIFAR10 classification task.}\label{steady}
\end{figure}

For our evaluation, first, the Accordion encoding is compared to the traditional encoding in means of convergence rate and overall accuracy. To do this, each encoding was used alongside the three individual GA training schemes for two network architectures each doing a classification task over a different dataset. From this comparison, we can also determine which scheme performs better than the others. Finally, the better encoding alongside the superior training scheme was compared to two backpropagation methods, Stochastic Gradient Descent and Adam training.\\
For the first part of the evaluation, the training methods were used for two network architectures each doing a classification task over the MNIST \cite{lecun-mnisthandwrittendigit-2010} and CIFAR10 \cite{krizhevsky2009learning} datasets, respectively. The architecture description of these networks are demonstrated in \autoref{tab:mnist} and \autoref{tab:cifar}. The network architecture for the MNIST task is custom design. But the network architecture for the CIFAR10 task is the famous LeNet network \cite{lecun2015lenet}.
\begin{table}[h]
\centering
\scriptsize
 \caption{\small The network architecture used for the MNIST classification task.}
\begin{tabular}{ccc}
\hline
Layer (type)                                                                                                                                     & Output shape         & Number of parameters \\ \hline
conv2d\_1 (Conv2D)                                                                                                                               & (None, 28, 28, 40)   & 160                  \\ \hline
max\_pooling2d\_1 (MaxPooling2D)                                                                                                                 & (None, 14, 14, 40)   & 0                    \\ \hline
conv2d\_2 (Conv2D)                                                                                                                               & (None, 14, 14, 40)   & 3200                 \\ \hline
max\_pooling2d\_2 (MaxPooling2D)                                                                                                                 & (None, 7, 7, 20)     & 0                    \\ \hline
conv2d\_3 (Conv2D)                                                                                                                               & (None, 7, 7, 5)      & 400                  \\ \hline
max\_pooling2d\_3 (MaxPooling2D)                                                                                                                 & (None, 3, 3, 5)      & 0                    \\ \hline
conv2d\_4 (Conv2D)                                                                                                                               & (None, 3, 3, 1)      & 20                   \\ \hline
flatten\_1 (Flatten)                                                                                                                             & (None, 9)            & 0                    \\ \hline
dense\_1 (Dense)                                                                                                                                 & (None, 40)           & 360                  \\ \hline
dense\_2 (Dense)                                                                                                                                 & (None, 10)           & 400                  \\ \hline
\multicolumn{1}{l}{\begin{tabular}[c]{@{}l@{}}Total parameters : 4540\\ Trainable parameters : 4540\\ Non-trainable parameters : 0\end{tabular}} & \multicolumn{1}{l}{} & \multicolumn{1}{l}{} \\ \hline
\end{tabular}
 \label{tab:mnist}
\end{table}
\begin{table}[h]
\centering
\scriptsize
 \caption{\small The network architecture used for the CIFAR10 classification task.}
\begin{tabular}{ccc}
\hline
Layer (type)                                                                                                                                         & Output shape         & Number of parameters \\ \hline
conv2d\_1 (Conv2D)                                                                                                                                   & (None, 32, 32, 6)    & 150                  \\ \hline
average\_pooling2d\_1 (AveragePooling2D)                                                                                                             & (None, 16, 16, 6)    & 0                    \\ \hline
conv2d\_2 (Conv2D)                                                                                                                                   & (None, 16, 16, 16)   & 2400                 \\ \hline
average\_pooling2d\_2 (AveragePooling2D)                                                                                                             & (None, 8, 8, 16)     & 0                    \\ \hline
conv2d\_3 (Conv2D)                                                                                                                                   & (None, 8, 8, 120)    & 48000                \\ \hline
average\_pooling2d\_3 (AveragePooling2D)                                                                                                             & (None, 4, 4, 120)    & 0                    \\ \hline
flatten\_1 (Flatten)                                                                                                                                 & (None, 1920)         & 0                    \\ \hline
dense\_1 (Dense)                                                                                                                                     & (None, 120)          & 230400               \\ \hline
dense\_2 (Dense)                                                                                                                                     & (None, 84)           & 10080                \\ \hline
dense\_3 (Dense)                                                                                                                                     & (None, 10)           & 840                  \\ \hline
\multicolumn{1}{l}{\begin{tabular}[c]{@{}l@{}}Total parameters : 291870\\ Trainable parameters : 291870\\ Non-trainable parameters : 0\end{tabular}} & \multicolumn{1}{l}{} & \multicolumn{1}{l}{} \\ \hline
\end{tabular}
 \label{tab:cifar}
\end{table}
\\For the MNIST classification task network, the folded chromosome has a length of 6, that is the 4 convolution filters conv2d\_1, ..., conv2d\_4 and the 2 dense layers dense\_1 and dense\_2. This small chromosome expands to a length of 136, that is 40 elements for the conv2d\_1 layer, 40 elements for the conv2d\_2 layer, 5 elements for the conv2d\_3 layer, 1 element for the conv2d\_4 layer, 40 elements for the dense\_1 layer and finally, 10 elements for the dense\_2 layer. For the CIFAR10 classification task network, again, the folded chromosome has a length of 6, that is the 3 convolution filters conv2d\_1, ..., conv2d\_3 and the 3 dense layers dense\_1, ..., dense\_3. This chromosome expands to a length of 356, that is 6 elements for the conv2d\_1 layer, 16 elements for the conv2d\_2 layer, 120 elements for the conv2d\_3 layer, 120 elements for the dense\_1 layer, 84 elements for the dense\_2 layer and finally, 10 elements for the dense\_3 layer.\\
In all of the used schemes, $pop\_size$ was set to 100. Meaning that 100 networks are initialized in the beginning.\\
For the steady-state scheme, an unfair coin-flip decision making is used, so that 70\% of the time, a child is produced using the crossover operator and the other 30\% using mutation. For the mutation operator in this scheme, the fully expanded chromosome is considered. Next, for each of the network's sections (feature extraction or classification), this operator selects $mutation\_ratio\times section\_length$ of elements of that section randomly and uniformly to be mutated according to \autoref{sec:steadyscheme}. The $mutation\_ratio$ was set to 0.1 for this scheme. For example, in case of the MNIST classification task network, that the fully expanded chromosome has a length of 136, its first 86 elements belong to the feature extraction section and the other 50 belong to the fully connected sections. So, for the first and the second sections, $int(86\times0.1)=8$ and $int(50\times0.1)=5$ elements will be selected to undergo mutation.\\
For the generational scheme, the mating pool has the same size as the population, that is 100 members. Then, its members are paired into 50 set of parents that each spawn two children, creating 100 new children. Next, some of these children will be mutated. A mutation probability of 0.2 is considered for the intermediate generation of size 100. Meaning that each one of them will have a chance of 0.2 to undergo mutation. The $mutation\_ratio$ and the operator's run description for this scheme is the same as the previous scheme.\\
Finally, for the elitism scheme, the top 10 best and fittest members of the population are moved directly to the next generation. This means that we need to create 90 children. For this, we select 20 members to enter the mating pool. Like before, we create and mutate 90 new members to enter into the next generation.\\
Now, the performance of each encoding used in the mentioned schemes for the two networks will be measured against their respective datasets. For the MNIST dataset, 70000 images are used for training and validation and 10000 images for testing. Also, for the CIFAR10 dataset, 50000 images are used for training and validation and 10000 images are used for testing. In the process of training all of the upcoming test cases, batch size is considered 32. So each iteration of the training algorithm inputs 32 images from the dataset and learns from that. Another important note here is that since the number of iterations for the following evaluation cases are so high, for the sake of plot visibility, the logarithm of the number of iterations is considered.\\
We must also note here a very important factor in the fidelity of our results, that is the initialization used for all of the training methods for each network is the same. Meaning that, for example, for the MNIST network, a population is initialized, and then, this one and only population is evolved using different encoding and training schemes. This results in the same starting accuracy point for every comparison that will be made. Also, when comparing to backpropagation methods later, the fittest member of the initialization population is selected to be trained with backpropagation.\\
At \autoref{fig:SteadyMNISTEncodingVS}, the network for the MNIST classification task was trained using the steady-state scheme using one time the traditional encoding and one time using the Accordion encoding. It can be derived from this result that the Accordion encoding performs only slightly better than the traditional encoding. This evaluation is also carried out for the CIFAR10 classification task. As shown in the results for this evaluation in \autoref{fig:StadyCIFAREncodingVS}, even though that the initial accuracy in the population of networks that were meant to be trained using the Accordion encoding is lower than of its same for the traditional encoding, the population trained using the Accordion encoding manages to surpass the traditional encoding multiple time in the evolution progress. And additionally, evolution using the Accordion encoding achieves a higher accuracy threshold compared to the traditional encoding in the same number of iterations. In short, the Accordion encoding has a faster convergence rate and reaches better accuracies against the traditional encoding.
\\Seeing in \autoref{fig:GenerationalMNISTEncodingVS} and \autoref{fig:GenerationalCIFAREncodingVS}, the same networks for the MNIST and CIFAR10 classification task were trained using the Accordion encoding and the traditional encoding with the generational scheme and their performance were measured against each other. We can clearly see that this scheme performs very oscillated, regardless of the encoding paradigm used. Nonetheless, it is obvious that the training using the Accordion encoding hits higher accuracies sooner in comparison with the traditional encoding. In means of overall convergence rate, both encoding paradigms do pretty much the same.
\\Finaly in \autoref{fig:ElitismMNISTEncodingVS} and \autoref{fig:ElitismCIFAREncodingVS}, the same network for the MNIST and CIFAR10 classification task were trained using the Accordion encoding and the traditional encoding with the elitism scheme. Comparing the performance of these two training methods, again, the Accordion encoding reaches higher accuracies in the same amount of iterations compared to the traditional encoding, proving its superiority. Moreover, the convergence rate of the two encoding paradigms are again pretty much the same.
\\
\\For the next part of our evaluation, we compare the performance the three training schemes for the two mentioned networks. All of these schemes employed the Accordion encoding. In \autoref{fig:ComulativeMNIST} and \autoref{fig:ComulativeCIFAR}, we can see the performance of these three schemes for training the two networks.
As depicted, the steady-state scheme delivers a slow but steady convergence evolution process. This scheme doesn't move that much from its start in means of accuracy. But because of its design, we can say for sure that the evolution process either improves or stays the same and it never gets worse. This scheme will evolve the population of solutions slow, but steady.
The generational scheme, while acting very oscillated, performs better than the steady-state scheme, as it reaches higher accuracies much faster. This scheme hits all sorts of accuracies in the duration of its evolution, allowing to score higher, faster than the steady-state scheme. But due to its oscillated nature, it does not stay there long enough. This scheme allows the population to explore a large area of the solution domain quite fast, but does not make an effort to keep them around one place much. To put it more technically, this scheme does a good job of exploring the solution domain for findings global optimum, but falls short in scaling it. In the means of overall convergence rate, this scheme does not differ that much with the stead-state scheme. They both eventually reach the same accuracy around the same time.
lastly, for the elitism scheme, we can see that our intuitions was right and this scheme delivers the best of both previous schemes. It has the stable and increasing rate of growth same as the steady state scheme while reaching higher accuracies. It has the same satisfying solution domain sampling of the generational scheme while having a stable converges that guarantees the best of a population does not get any worse. It is clear that passing the elite members directly into the next generation plays a key role in the good performance of this scheme.
\\Finally, for the last part of our evaluation, the best encoding, that was proved to be the Accordion encoding, used in the Superior training scheme, that was concluded to be elitism, are used to train the two mentioned networks and be compared with training the same networks using backpropagation training with Adam optimizer.\\
As shown in \autoref{fig:MNIST}, the GA training reaches lower accuracies much faster than Adam training. But it eventually falls short drastically. However, as shown in \autoref{fig:CIFAR}, the GA training performs tremendously better than Adam training. Intuition here suggest that as the network size grows and the task gets more complicated (as of CIFAR10 over MNIST), Adam training does not scale the solution domain as fast as the GA training samples it in the early stages.
\\To summarize our evaluation, we first compared the Accordion encoding with the traditional encoding to find out which one was better. Our results clearly demonstrated the effectiveness of the Accordion encoding. Next, we evaluated the three GA training schemes against one another, proving the superiority of the elitism scheme. And finally, The best encoding and scheme was compared to backpropagation, showing the promising property of the GA training methods to scale the solution domain faster in the early stages of the training process.
\section{Conclusion}\label{sec:conclusion}
In this work, we proposed novel methods for training deep convolutional neural networks based on genetic algorithm. Our proposed methodology involves novel genetic operators for crossover and mutation and a new encoding paradigm of the networks to the chromosomes. This new encoding paradigm delivers a significant reduction in the chromosome size, by considering the entirety of a layer (weather its convolution or fully connected) as genes in the chromosome. These methodologies are carried out using three different schemes of genetic algorithm, steady-state, generational, and elitism. We first evaluated the performance of our novel Accordion encoding against the traditional encoding. Our results demonstrated that the Accordion encoding performs better than the traditional encoding in means of convergence rate and overall accuracy. Next, the individual GA training schemes were compared to each other in means of convergence time and overall accuracy. We concluded that the steady-state scheme delivers a slow but steady convergence rate, the generational scheme performs much faster, but behaves oscillated, and the elitism scheme, expectedly, as a combination of both mentioned schemes, archives better accuracies in a much faster and more stable rate than the two other schemes. Finally, the Accordion encoding used in the superior training scheme, that is the elitism scheme, was compared to backpropagation training with Adam optimizer. We derived that even though GA training falls short in overall, it performs better in the early stages of the training, specially as the problem gets more complicated. This delivers a promising property to head-start the training process with genetic algorithm.\\



\end{document}